\documentclass[10pt,twocolumn,letterpaper]{article}

\pdfpageattr{/Group <</S /Transparency /I true /CS /DeviceRGB>>}

\usepackage{authblk}
\usepackage{varwidth}

\makeatletter
\renewcommand\AB@affilsepx{\hspace{1.2cm}}
\makeatother

\usepackage[dvipsnames,svgnames,x11names,table, dvipsnames]{xcolor}
\usepackage[markup=underlined]{changes}
\definecolor{blueNote}{HTML}{2B619D}
\definechangesauthor[color=blueNote,name=\textbf{Alejandro}]{AC}
\definechangesauthor[color=red,name=\textbf{Mariella}]{MD}
\usepackage{todonotes}
\makeatletter
\setremarkmarkup{\todo[color=Changes@Color#1!20,size=\scriptsize]{#1: #2}}
\makeatother

\usepackage{iccv}

\usepackage{times}
\usepackage{amsmath}
\usepackage{amssymb}

\usepackage{multirow}

\usepackage{epsfig}
\usepackage{graphicx}
\usepackage[export]{adjustbox}

\usepackage{array}
\usepackage{comment}
\newcolumntype{R}[2]{%
    >{\adjustbox{angle=#1,max width=1.5cm,lap=\width-(#2)}\bgroup}%
    l%
    <{\egroup}%
}
\newcommand*\rot{\multicolumn{1}{R{90}{1em}}}

\usepackage{amsmath,amssymb} 

\usepackage[pagebackref=true,breaklinks=true,letterpaper=true,colorlinks,bookmarks=false]{hyperref}
\usepackage{subcaption}
\iccvfinalcopy 

\def\iccvPaperID{} 

\ificcvfinal\fi

\makeatletter
\def\@maketitle{%
  \newpage
  \vspace{-0.5cm}
  \begin{center}%
  \let \footnote \thanks
    {\Large\textbf {\@title} \par}%
    \vskip 1.5em%
    {\large
      \lineskip .5em%
      \begin{tabular}[t]{c}%
        \@author
      \end{tabular}\par}%
    \vskip 1em%
  \end{center}%
  \par
  \vskip 1.5em}
\makeatother

\begin{document}

\title{On the Role of Event Boundaries in Egocentric Activity Recognition from Photostreams}

\author[1,2]{Alejandro Cartas}
\author[1]{Estefania Talavera}
\author[1,2]{Petia Radeva}
\author[1,2]{Mariella Dimiccoli}

\affil[1]{\hspace{-2.1cm}\protect\begin{varwidth}[t]{\linewidth}\protect\centering University of Barcelona\par Mathematics and Computer Science Department\par 08007 Barcelona\par Spain \authorcr {\tt\small \{alejandro.cartas, etalavera, petia.ivanova\}@ub.edu} \protect\end{varwidth}}
\affil[2]{\hspace{-0.85cm}\protect\begin{varwidth}[t]{\linewidth}\protect\centering Computer Vision Center\par Universitat Aut\'onoma de Barcelona\par 08193 Cerdanyola del Vallès\par Spain \authorcr {\tt\small mariella.dimiccoli@cvc.uab.es} 
\protect\end{varwidth}}

\maketitle

\begin{abstract}

Event boundaries play a crucial role as a pre-processing step for detection, localization, and recognition tasks of human activities in videos. Typically, although their intrinsic subjectiveness, temporal bounds are provided manually as input for training action recognition algorithms. However, their role for activity recognition in the domain of egocentric photostreams has been so far neglected. In this paper, we provide insights of how automatically computed boundaries can impact activity recognition results in the emerging domain of egocentric photostreams. Furthermore, we collected a new annotated dataset acquired by 15 people by a wearable photo-camera and we used it to show the generalization capabilities of several deep learning based architectures to unseen users. 
\end{abstract}

\section{Introduction}
\label{sec:introduction}

Wearable cameras offer a hand-free way to capture the world from a first-person perspective, hence providing rich contextual information about the activities being performed by the user \cite{Nguyen2016}. Similarly to other wearable sensors, wearable cameras are ubiquitous and allow to capture daily activities in natural settings.

\begin{figure}[t]
\centering
\includegraphics[width=\columnwidth]{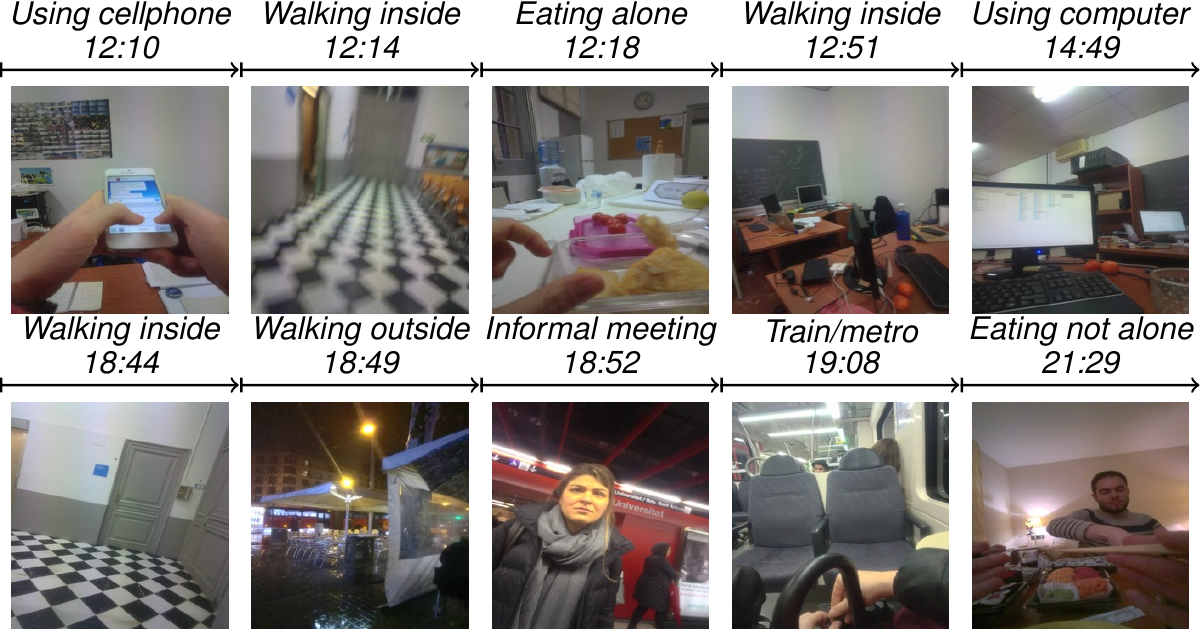}
\caption{Sample images captured by a wearable photo-camera user during a day, together with their timestamp and activity label.}
\label{fig:eventsPhotoStreams}
\end{figure}

Currently, recognizing daily activities from first-person (egocentric) images and videos is a very active area of research in computer vision \cite{Ma_2016_CVPR,poleg2016compact,castro2015predicting,cartas2017batch,cartas2017recognizing}. In this paper, we focus on streams of images captured at regular intervals through a wearable photo-camera, also called \textit{photostreams}, that have received comparatively little attention in the literature. With respect to egocentric videos, photostreams usually cover the full day of a person (see Fig. \ref{fig:eventsPhotoStreams}). However, since the photo-camera typically takes a picture every 30 seconds, temporally adjacent images present abrupt changes. Consequently, optical flow cannot be reliably estimated and several fine-grained action are completely missed or too sampled for being identifiable. Since motion is an important feature to disambiguate activities, recognize them become particularly challenging in the photostream domain. 

\begin{figure*}[t]
\centering
\adjincludegraphics[width=0.9\textwidth,trim={0 3cm 0 0cm},clip]{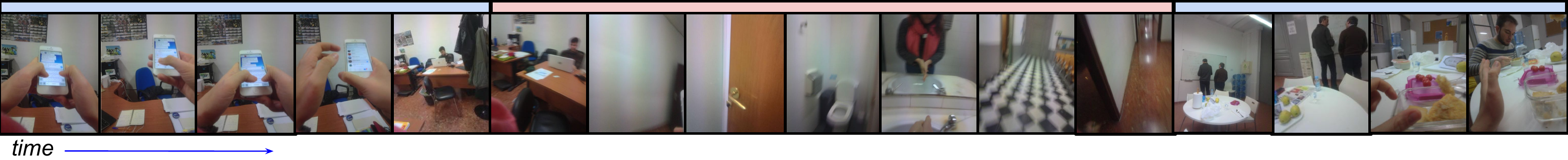}
\caption{Example of events obtained by applying SR-Clustering on a visual lifelog. The color above the images indicate correspondence to the event to which consecutive images belong.}
\label{fig:srclustSeg}
\end{figure*}

\begin{figure}[t]
\begin{center}
\includegraphics[width=0.08\textwidth]{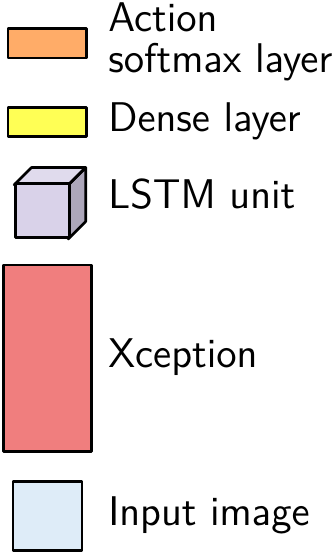}
\includegraphics[width=0.32\textwidth]{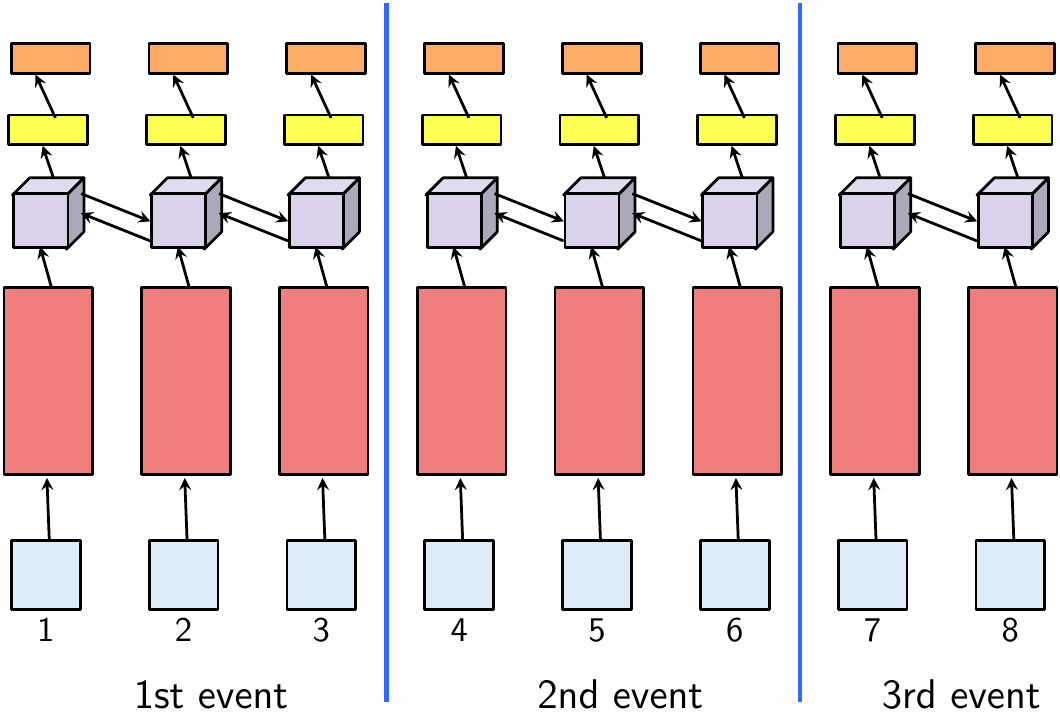}
\end{center}
\caption{Pipeline of our proposed approach.}
\label{fig:pipeline}
\end{figure}

Recently, several papers have proposed different deep learning architectures to recognize activities from egocentric photostreams. The earliest works \cite{castro2015predicting,cartas2017recognizing} focused on image-based approach, aimed at classifying each image independently by its neighbor frames. With the goal of taking advantage of the temporal coherence of objects that characterizes photostreams \cite{byrne2010everyday}, instead of working at image level, Cartas et al. \cite{cartas2017batch,Cartas2018} proposed to train in an end-to-end fashion a Long Short Term Memory (LSTM) recurrent neural network on the top of a CNN by feeding the LSTM using a sliding window approach. This strategy allows to copy with both the not negligible length of photostreams and the lack of knowledge of event boundaries. 

\begin{figure*}[t]
\centering
\includegraphics[width=0.75\textwidth]{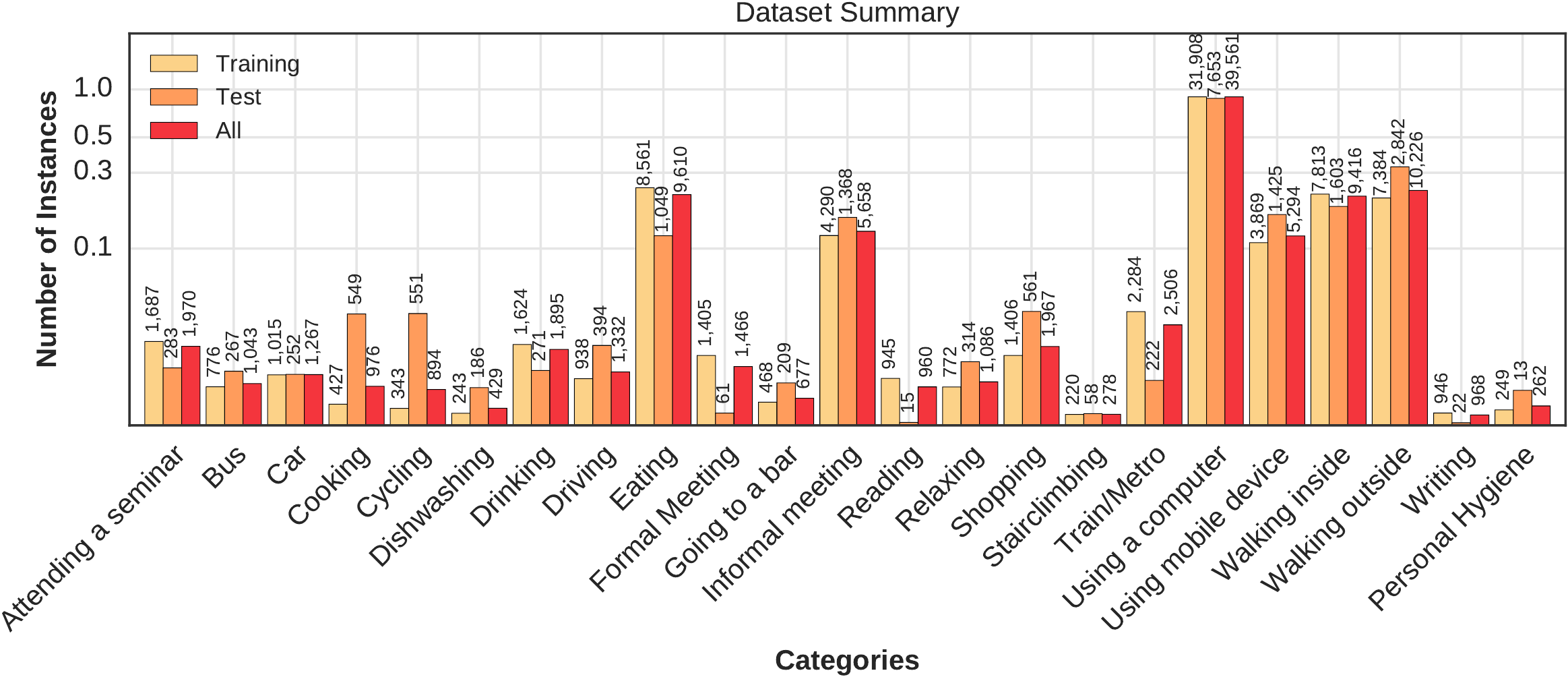}
\caption{Dataset summary. Please notice that distributions are normalized and the vertical axis has a logarithm scale.}
\label{fig:datasetSummary}
\end{figure*}

\begin{figure}[t]
\centering
\includegraphics[width=\columnwidth]{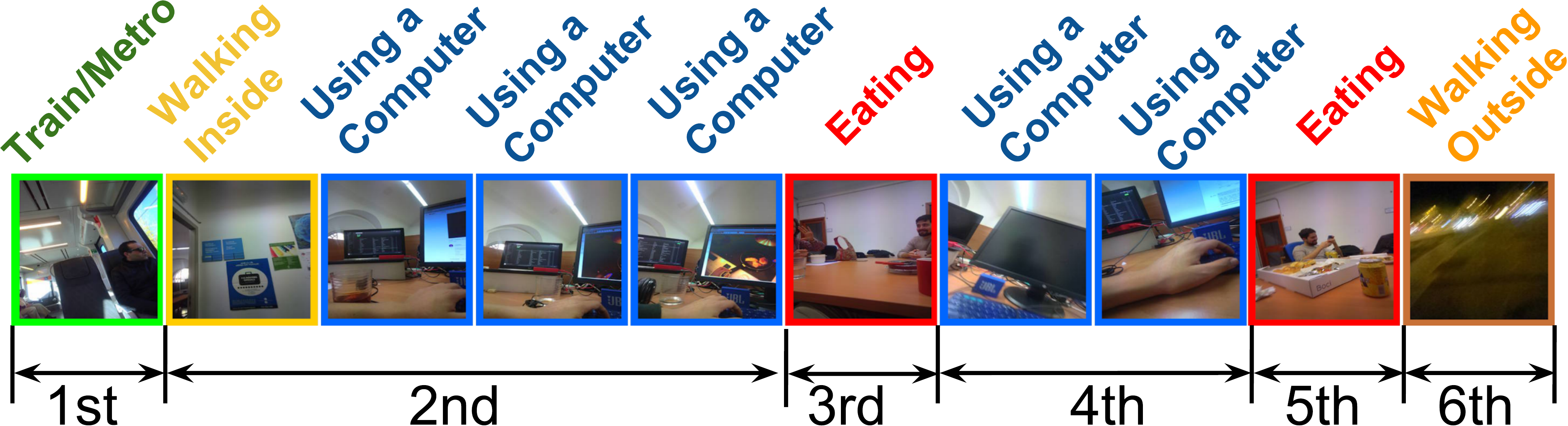}
\caption{Example of automatically extracted events used in the experiments.}
\label{fig:temporal_segments}
\end{figure}

This approach has showed that considering overlapping segments of fixed size turn out to be effective to better capture long-term temporal dependencies in photo-streams. In this paper,we argue that knowing exactly event boundaries would allow to further improve activity recognition performance, since it would allow to capture temporal dependencies both within an event and across events. 

\section{Event boundaries for activity recognition}
\label{sec:proposedApproach}

In this work, we investigate whether the use of event boundaries as additional input can improve the recognition of activities in egocentric photo-sequences. To this goal, we used the temporal segmentation method introduced in \cite{dimiccoli2017sr} that allows to extract events from long unstructured photostreams. Events obtained with such approach, correspond to temporally adjacent images the share both contextual and semantic features, as shown in Fig. \ref{fig:srclustSeg}. As it can be observed, these events constitute a good basis for activity recognition, since typically, when the user is engaged in an activity, contextual and semantic features have little variation. 

\section{Experimental setup}
\label{sec:experimentalSetup}

The objective of our experiments was to determine if the temporal coherence of segmented events from egocentric photostreams improved the activity recognition at the frame level. Therefore, we trained three many-to-many LSTM models using the full-day sequence and the automatically extracted event segments, i.e. CNN+RF+LSTM, CNN+LSTM, and CNN+Bidirectional LSTM (see Fig. \ref{fig:pipeline}) . For comparative purposes, we used as a baseline to train all models the Xception network \cite{Xception2017}. Additionally, we implemented the best model presented in \cite{Cartas2018}, namely the combination of CNN+RF+LSTM. We measure the activity recognition performance using the classification accuracy and associated macro metrics.\\

\textbf{Dataset}. We collected over 102,227 pictures from 15 college students who were asked to wear an egocentric camera\footnote{http://getnarrative.com/} on their chest. The camera automatically captured an image at $\approx30$ seconds rate with a 5MP resolution. The annotation process took into account the continuous context of activity sequences. In order to split the data in training and test sets, all the possible combinations of users for both sets were calculated. Only the combinations with a test set having all the categories and 20-21\% of all images were kept. A histogram of the number of photos per category and split is shown in Fig. \ref{fig:datasetSummary}. 

\textbf{Temporal sequences}. The following temporal sequences were used in the experiments:
\begin{enumerate}
\item \textit{Fixed size segments}. The stateful sliding window training procedure for fixed size segments from \cite{carta2017batch} for LSTM was also implemented.

\item \textit{Full sequence}. The whole day photostream sequence of each user were used as a single input.

\item \textit{Event segmentation}. Groups of sequential images were obtained by applying the method introduced by Dimiccoli et al. \cite{dimiccoli2017sr}, which temporally segments the given photostream as illustrated in Fig \ref{fig:temporal_segments}.
\end{enumerate}

\setlength{\tabcolsep}{4pt}
\begin{table*}[t!]
\begin{center}
\caption{Activity classification performance. Upper part shows the recall for each category and the lower part shows the performance metrics for all models. The best result per measure is shown in bold but does not take into account the temporal models trained using the groundtruth segmentation, that we consider as an upper bound.}
\label{table:activityClassificationPerformance}
\resizebox{0.7\textwidth}{!}{%
\begin{tabular}{ l | c |c !{\color{RoyalBlue}\vrule width 2pt} c |c |c !{\color{RoyalBlue}\vrule width 2pt} c |c |c !{\color{RoyalBlue}\vrule width 2pt}c |c |c }
& %
\multicolumn{2}{c}{}{\color{RoyalBlue}\vrule width 2pt}& %
\multicolumn{3}{c}{}{\color{RoyalBlue}\vrule width 2pt}& %
\multicolumn{3}{c}{}{\color{RoyalBlue}\vrule width 2pt}& %
\multicolumn{3}{c}{}%
\\
& %
\multicolumn{2}{c}{}{\color{RoyalBlue}\vrule width 2pt}& %
\multicolumn{3}{c}{\textbf{Xception+RF+LSTM}}{\color{RoyalBlue}\vrule width 2pt}& %
\multicolumn{3}{c}{\textbf{Xception+LSTM}}{\color{RoyalBlue}\vrule width 2pt}& %
\multicolumn{3}{c}{\textbf{Xception+Bidi LSTM}}%
\\
\textbf{Activity} & %
\rot{\parbox{1.2cm}{\textbf{Xception}}} & %
\rot{\parbox{1.2cm}{\textbf{Xception+RF}}} {\color{RoyalBlue}\vrule width 2pt}&  %
\rot{\parbox{2.0cm}{\textbf{Fixed size\\ segments}}} & %
\rot{\parbox{1.2cm}{\textbf{Full\\ sequence}}} & %
\rot{\parbox{2.0cm}{\textbf{Event\\ segmentation}}} {\color{RoyalBlue}\vrule width 2pt}& %
\rot{\parbox{2.0cm}{\textbf{Fixed size\\ segments}}} & %
\rot{\parbox{1.2cm}{\textbf{Full\\ sequence}}} & %
\rot{\parbox{2.0cm}{\textbf{Event\\ segmentation}}} {\color{RoyalBlue}\vrule width 2pt}& %
\rot{\parbox{2.0cm}{\textbf{Fixed size\\ segments}}} & %
\rot{\parbox{1.2cm}{\textbf{Full\\ sequence}}} & %
\rot{\parbox{2.2cm}{\textbf{Event\\ segmentation}}} \\ \hline 
\textbf{Accuracy} & 68.88 & 70.77 & 70.64 & 72.21 & 72.52 & 70.24 & 74.27 & 73.28 & 74.20 & 75.59 & \textbf{76.09} \\ \hline 
\textbf{Macro precision} & 52.04 & 52.02 & 35.16 & 42.71 & 54.81 & 48.06 & 59.03 & 57.71 & 51.95 & 56.81 & \textbf{59.29} \\ \hline 
\textbf{Macro recall} & 38.17 & 33.32 & 34.62 & 36.11 & 36.98 & 41.07 & 50.71 & 49.75 & \textbf{54.68} & 48.30 & 50.22 \\ \hline 
\textbf{Macro F1-score} & 39.05 & 32.23 & 32.71 & 35.44 & 36.44 & 40.19 & 50.85 & 48.94 & 50.66 & 48.50 & \textbf{51.21} \\ \hline
\end{tabular}

}
\end{center}
\end{table*}
\setlength{\tabcolsep}{1.4pt}

\setlength{\tabcolsep}{4pt}
\begin{table*}[t!]
\begin{center}
\caption{Comparison with different Egocentric datasets. Information based on \cite{Damen2018EPICKITCHENS}}
\label{table:datasetsComparison}
\resizebox{0.8\textwidth}{!}{%
\begin{tabular}{|l|c|c|c|c|r|c|r|c|c|}
\hline
 & &\textbf{Non-} &\textbf{Native} & & &\textbf{Action} & \textbf{Action/Activity} & \\
\textbf{Dataset} & \textbf{Photo-streams} &\textbf{Scripted?} &\textbf{Env?} &\textbf{Frames} & \textbf{Sequences} & \textbf{Segments} &\textbf{Classes}  & \textbf{Participants} \\
\hline
Ours &$\checkmark$ &$\checkmark$ &$\checkmark$ &0.1M  & 191 days  &-  &23 &15 \\
UT-Egocentric\cite{lee2012} & $\times$ & $\checkmark$ &$\checkmark$ &0.9M& 4  &-  &- &4 \\
KrishnaCam\cite{krishna-wacv2016} & $\times$ & $\checkmark$ &$\checkmark$ &7.6M& 460  &-  &- &1 \\
DECADE\cite{ehsani2018dog} & $\times$ & $\checkmark$ &$\checkmark$ &0.02M& 380  &-  &48 &1 \\
ADL\cite{Pirsiavash2012} &$\times$ &$\times$ &$\checkmark$ &1.0M  &20  &436  &32 &20 \\
Epic Kitchens\cite{Damen2018EPICKITCHENS} &$\times$ &$\checkmark$ &$\checkmark$ &11.5M  &432  &39,596 &149*  &32 \\
GTEA Gaze+\cite{Fathi2012}  &$\times$ &$\times$ &$\times$ & 0.4M &35  &3,371  &42 &13 \\
CMU\cite{DeLaTorre2008} &$\times$ &$\times$ &$\times$ & 0.2M &16  &516 &31 &16 \\
BEOID\cite{Damen2014a} &$\times$ &$\times$ &$\times$ &0.1M &58  &742  &34 &5\\
\hline
\end{tabular}
}
\end{center}
\end{table*}
\setlength{\tabcolsep}{1.4pt}

\section{Experimental results}
\label{sec:results}

In Table \ref{table:activityClassificationPerformance} we present the performance of all the models using full sequence, SR-Clustering (event segmentation), and the sliding window training procedure (fixed size segments) proposed in \cite{cartas2017batch}. The performance was evaluated using the accuracy and macro metrics for precision, recall, and F1-score.

The results indicate that the CNN+Bidirectional LSTM model achieves the best performance over all the models and on each temporal segmentations. On the other hand, the CNN+RF+LSTM model did not improved the performance as much as the other models and was even worse than its baseline using the sliding window training. This is a consequence of the overfitting of its base model (CNN+RF) in the training set, as shown by the categories recall in Table \ref{table:activityClassificationPerformance}. This contrasts the results previously obtained in \cite{Cartas2018} using another dataset and it is likely due to the fact that here we are using non-seen users in our test set.

\begin{figure}[t!]
\centering
\includegraphics[width=\columnwidth]{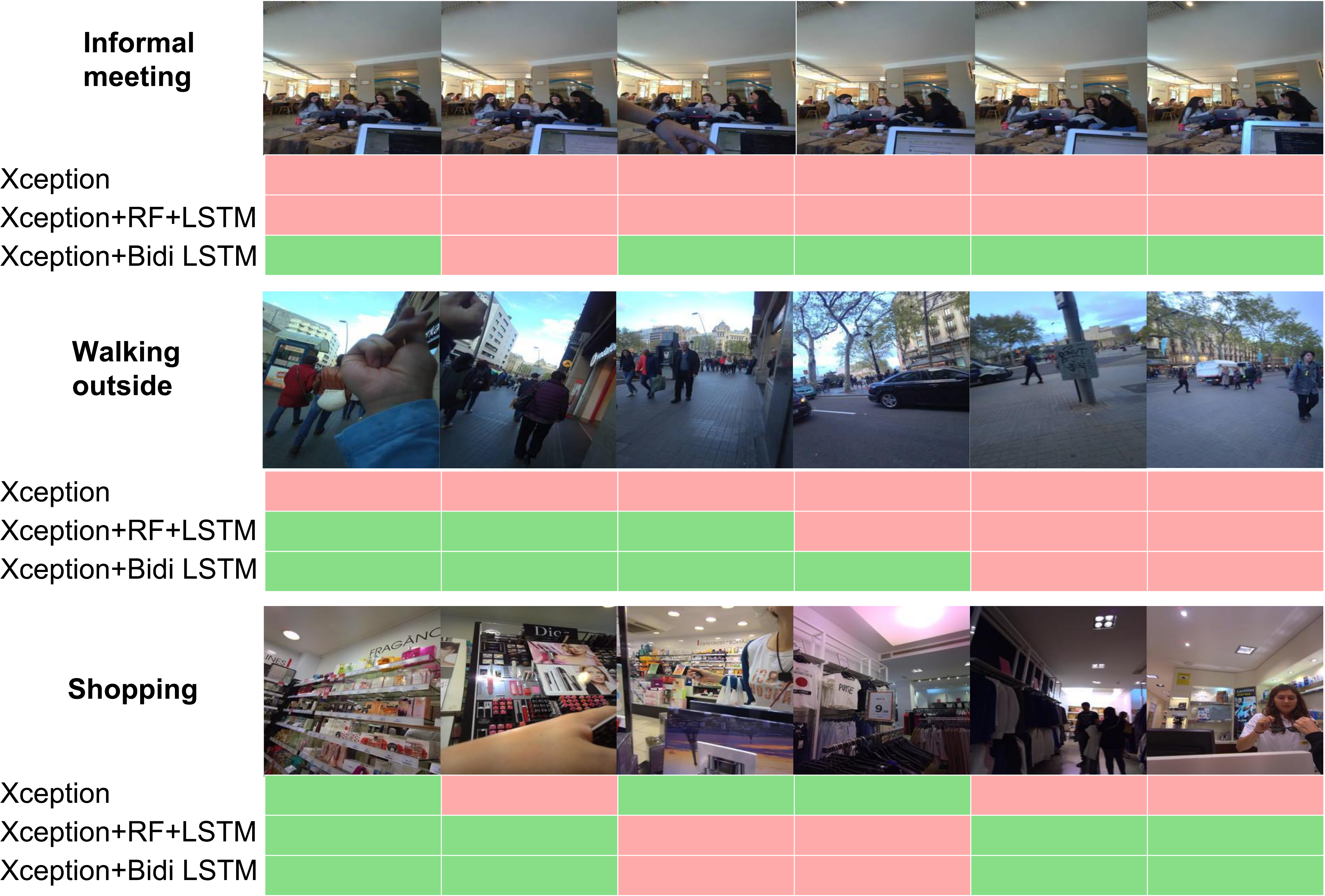}
\caption{Examples of qualitative results obtained from three of the evaluated methods (Xception, Xception+RF+LSTM, and Xception+Bidirectional LSTM) for different activity classes. False and true activity labels for a given image are marked in red and green, respectively.}
\label{fig:qualitatresults}
\end{figure}

Furthermore, the results suggests that the temporal segmentation increased the classification performance of the tested LSTM based models. For instance, Fig. \ref{fig:qualitatresults} shows some qualitative results. In particular, the automatic event segmentation (SR-Clustering) was better than the day segmentation as it improved the accuracy, macro precision, and macro F1-scores in two of the three LSTM based models. Since most of the test users had short day sequences, the day temporal segmentation was the best for CNN+LSTM model. Finally, the best macro recall was obtained using the Sliding Window training \cite{Cartas2018} for the CNN+Bidirectional LSTM model. This can be understood as a smoothing effect over the test sequences.

\section{Conclusions}
\label{sec:conclusions} 

This paper has shed light on two poorly investigated issues in the context of activity recognition from egocentric photostreams. The first issue was related to the role of event boundaries as input for activity recognition in photostreams. By relying on manually-annotated and automatically-extracted event boundaries, in addition to overlapping batches of images of fixed size, this paper pointed out that activity recognition performances benefit from the knowledge of event boundaries. The second issue was related to the generalization capabilities of existing methods for activity recognition. By using a large egocentric dataset acquired from 15 users, this paper could elucidated for the first time, how activity recognition performance generalize at test time to unseen users. The best results were achieved by using a CNN+Bidirectional LSTM architecture on a temporal event segmentation.

\section*{Acknowledgments}

A.C. was supported by a doctoral fellowship from the Mexican Council of Science and Technology (CONACYT) (grant-no. 366596). This work was partially founded by TIN2015-66951-C2, SGR 1219, CERCA, \textit{ICREA Academia'2014} and 20141510 (Marat\'{o} TV3). The funders had no role in the study design, data collection, analysis, and preparation of the manuscript. M.D. is grateful to the NVIDIA donation program for its support with GPU card.

{\small
\bibliographystyle{ieee}

}
\end{document}